\documentclass{article}

\usepackage{arxiv}

\usepackage[utf8]{inputenc} % allow utf-8 input
\usepackage[T1]{fontenc}    % use 8-bit T1 fonts
\usepackage{hyperref}       % hyperlinks
\usepackage{url}            % simple URL typesetting
\usepackage{booktabs}       % professional-quality tables
\usepackage{amsfonts}       % blackboard math symbols
\usepackage{nicefrac}       % compact symbols for 1/2, etc.
\usepackage{microtype}      % microtypography
\usepackage{lipsum}
\usepackage{graphicx}
\usepackage{wrapfig}
\usepackage[english]{babel}
\usepackage[nottoc]{tocbibind}
\usepackage{float}
\usepackage{caption}

\graphicspath{ {./Figures/} }

\title{Training Set Effect on Super Resolution for Automated Target Recognition}

\begin{document}
\maketitle

\begin{table}[H]
\vspace{-100pt}
\centering
Matthew Ciolino, David Noever, Josh Kalin
\vspace{15pt}
\caption*{PeopleTec, Inc. -- Huntsville, AL 35805 -- 256-319-3800}
\vspace{80pt}
\end{table}

% \begin{table}[H]
% \vspace{-100pt}
% \centering
% \begin{tabular}{l|l}
% Matthew Ciolino & matt.ciolino@peopletec.com      \\
% David Noever & david.noever@peopletec.com      \\
% Josh Kalin & josh.kalin@peopletec.com        \\
% Dominick Hambrick & dominick.hambrick@peopletec.com \\
% \end{tabular}
% \vspace{15pt}
% \caption*{PeopleTec, Inc. -- Huntsville, AL 35805 -- 256-319-3800}
% \vspace{80pt}
% \end{table}

\vspace{-65pt}

\begin{abstract}
Single Image Super Resolution (SISR) is the process of mapping a low-resolution image to a high resolution image. This inherently has applications in remote sensing as a way to increase the spatial resolution in satellite imagery. This suggests a possible improvement to automated target recognition in image classification and object detection. We explore the effect that different training sets have on SISR with the network, Super Resolution Generative Adversarial Network (SRGAN). We train 5 SRGANs on different land-use classes (e.g. agriculture, cities, ports) and test them on the same unseen dataset. We attempt to find the qualitative and quantitative differences in SISR, binary classification, and object detection performance. We find that curated training sets that contain objects in the test ontology perform better on both computer vision tasks while having a complex distribution of images allows object detection models to perform better. However, Super Resolution (SR) might not be beneficial to certain problems and will see a diminishing amount of returns for datasets that are closer to being solved.

% We find that curated training sets that contain objects in the test ontology perform better in the computer vision tasks. However, Super Resolution (SR) might not be beneficial to certain problems and will see a diminishing amount of returns for data sets that are closer to being solved.

%Do we need to have more expensive remote sensing satellites when we could use single image super-resolution (SISR) to get the spatial resolution that we want? By using a Super Resolution Generative Adversarial Network, (SRGAN) we can get higher resolution images. Previous work by Shermeyer et al. \cite{shermeyer2019effects} have used SISR as a preprocessing step describe an increase in mAP of 10-36 \% in object detection for native 30cm to 15cm satellite imagery. This suggests a possible improvement to automated target recognition in image classification and object detection. The SRGAN takes a low-resolution image and maps it to a high-resolution image creating the super resolution product. We train 5 SRGANs on different land-use classes (e.g. agriculture, cities) and find the qualitative and quantitative differences in SISR, binary classification, and object detection performance.

\end{abstract}

\begin{wrapfigure}{r}{0.45\textwidth}
    \vspace{-40pt}
    \begin{center}
        \includegraphics[width=.45\textwidth]{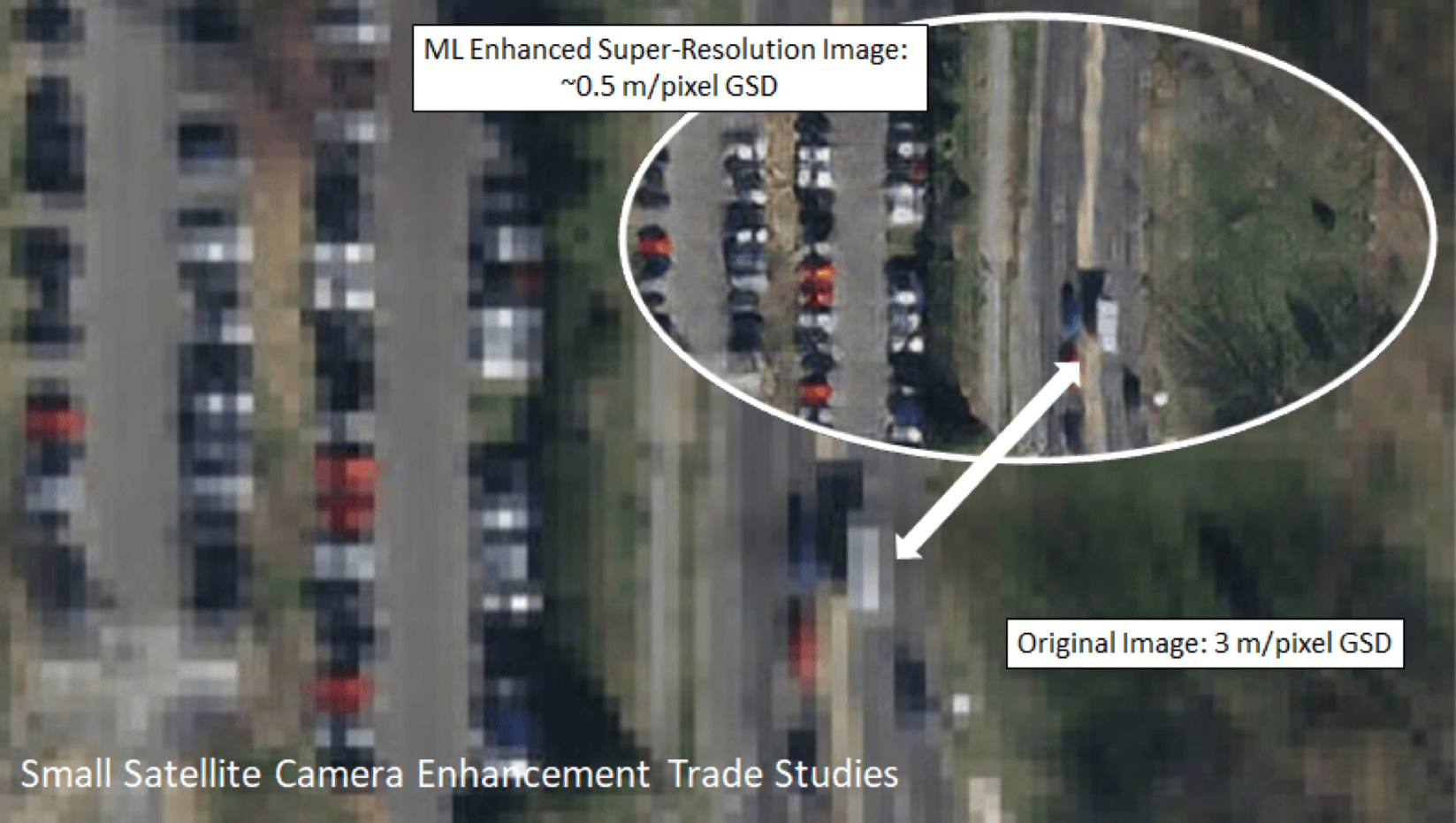}
    \end{center}
    \caption{SR's effect of increasing spatial resolution}
    \vspace{40pt}
    \label{fig:1}
\end{wrapfigure}

\section{Introduction}
\label{sec:headings}
Single Image Super Resolution is the process of taking a low resolution (LR) image and running it through a model to increase the resolution with higher fidelity information than any scaling algorithm (Fig \ref{fig:1}). This process currently does and has the potential to remove the need for increasingly large and expensive satellite cameras as running SISR could effectively increase the spatial resolution of your images. Since there are a multitude of ways to increase the resolution of an image, this is an ill-posed problem with many possible solutions. While significant work has been done on non-satellite images for SISR, not a lot has been done for satellite specific SR networks. In addition, most papers have tried to show the improvement in model scores while the purpose of this paper is to show the difference in how the networks are trained and its effect on computer vision tasks.

\begin{wrapfigure}[5]{r}{0.45\textwidth}
    \vspace{-80pt}
    \begin{center}
        \includegraphics[width=.45\textwidth]{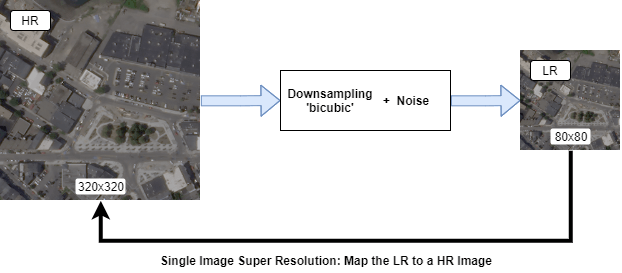}
    \end{center}
    \caption{Overview of SISR}
    \vspace{80pt}
    \label{fig:2}
\end{wrapfigure}

\subsection{Past Works}
A comprehensive review of SISR can be found here \cite{yang2019deep}, but we attempt to provide a brief summary. SISR at its core is trying to map a low-resolution image to a higher resolution image based upon the pixels in the image (Fig \ref{fig:2}). This process has had a growing interest over past years due to the rise of deep learning and the growth of computing power. Yang et al. states that there are three categories of SISR: interpolation-based, reconstruction-based, and learning-based methods. ‘Bicubic’ \cite{keys1981cubic} is the most popular algorithm for interpolation-based methods. An example of reconstruction-based method is \cite{dai2009softcuts}  where Dai et al. details a soft edge smoothness algorithm. The final category is learning based methods which is the realm for this paper.

\subsection{Related Works}
A few papers have sparked our interest in performing this experiment. In \cite{kawulok2019training}, Kawulok et al. tested the down sampling method for training SR networks. They tested SRResNet and FSRCNN using DIV2K images \cite{agustsson2017ntire} and Sentinal-2 images \cite{sentinel2a} (10m). They showed different image quality scores from different down sampling methods: Nearest Neighbors (NN), Bilinear, Bicubic, Lanczos, Lanczos-B, Lanczos-N, Lanczos-BN, and Mixed. NN showed the highest quality with ~31 PSNR with other methods mostly scoring around ~28 PSNR. \\
\phantom{x}\hspace{3ex} Furthermore, in \cite{shermeyer2019effects}, Shermeyer et al. performed SR as preprocessing step for object detection. The SR method used was Very-Deep-Super- Resolution \cite{kim2016accurate} (VDSR) which was introduced in 2016 and used residual-learning and extremely high learning rates to optimize a very deep network fast. They scored the two object detection methods, YOLT \cite{van2018you} and SSD \cite{liu2016ssd} and compared mean average precision (mAP) before and after super resolution. Shermeyer et al. found that for YOLT and SSD the largest gain in performance is achieved at the highest resolutions, as super-resolving native 30 cm imagery to 15 cm yields a 13-36\% improvement in mAP. \\
\phantom{x}\hspace{3ex} In our research, we found another paper looking at the difference in performance from SRGAN training sets on non-satellite imagery \cite{takano2019srgan}. They used the same network as this paper and applied it to datasets: CelebA \cite{liu2018large}, Dining Room \cite{yu2015lsun}, and Tower \cite{yu2015lsun}. Intuitively, Takano et al. found that test images that contained objects from the training set performed much better than test images that contained no objects from the training set.\\
\phantom{x}\hspace{3ex} These papers and the changes they made to the single image super resolution pipeline was the inspiration for studying the effects that different training sets have on super resolution. We will first look at the networks used in the experiment before diving into the datasets used and the experiment itself. 

\section{Networks}
\label{sec:headings}
This experiment used 3 networks; SRGAN for SR, a CNN for image classification and Mask R-CNN for object detection. 

\begin{wrapfigure}[16]{r}{0.25\textwidth}
    \vspace{-40pt}
    \begin{center}
        \includegraphics[scale=.45]{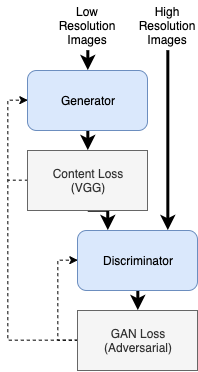}
    \end{center}
    \caption{GAN flow outline}
    \vspace{40pt}
    \label{fig:3}
\end{wrapfigure}

\subsection{SRGAN}
\begin{figure}[h]
\includegraphics[scale=.195]{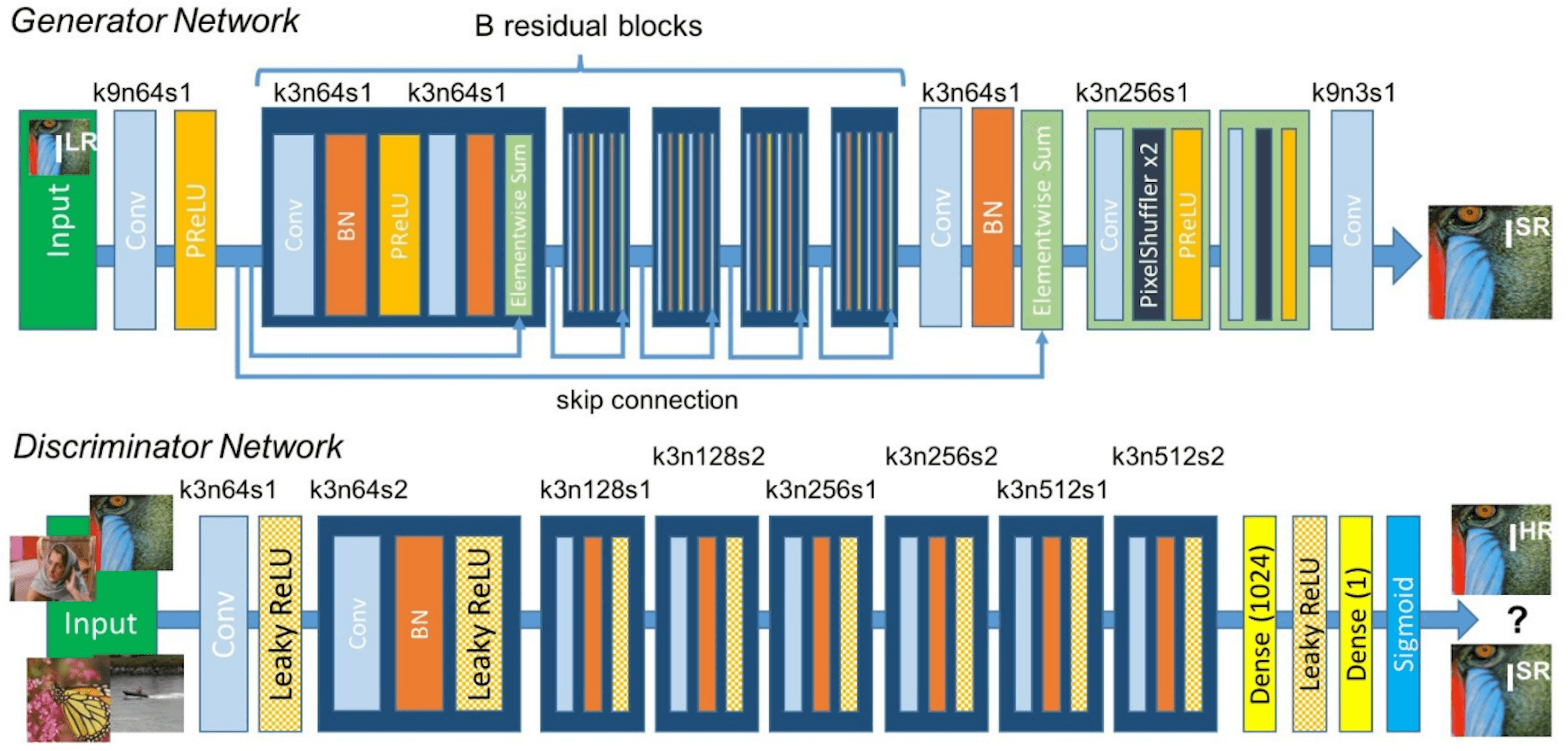}
\centering
\caption{Architecture of Generator and Discriminator Network with corresponding kernel size (k), number of feature maps (n) and stride (s) indicated for each convolutional layer. Copy of original network image in \cite{ledig2017photo}.}
\label{fig:4}
\end{figure}

We use the SRGAN as described in \cite{ledig2017photo} by Ledig et al in 2017. This was implemented in Keras in \cite{deepak112_2019} with minor changes. The GAN in general flows from a generator to a discriminator (Fig \ref{fig:3}).  We down sample a high-resolution (HR) image into a LR image. Then we take the LR images and pass them through the generator to get SR images. We then compare the HR image to the SR image to get the content loss. We then pass the HR and SR images to the discriminator to try to predict if it is a fake image or not. We then get the GAN loss and pass that back to the generator. The generator and the discriminator networks can be seen in detail here (Fig \ref{fig:4}) as shown in the original paper by Ledig. The interesting additions to the GAN network are the loss functions used \cite{ledig2017photo}. The loss function for the discriminator is a Keras default binary cross entropy while the loss function for the generator (Eq \ref{eq:3}) consists of two parts, content loss (Eq \ref{eq:2}) and adversarial loss (Eq \ref{eq:1}). As Ledig et al. describes, at large upscaling factors pixel-wise loss, such as mean squared error, fails to capture high-frequency content and leads to smoothed textures. Therefore, the content loss used is a perceptual loss function which compares the weights of the 19th layer of the VGG19 network for the HR vs SR images. As stated by Ledig et al, this was used in \cite{johnson2016perceptual} for style transfer. 

\begin{equation}
    l^{SR}_{GEN}=\sum_{n=1}^n -log( D_{\theta_D}(G_{\theta_G}(I^{SR})))
    \label{eq:1}
\end{equation}

\begin{equation}
    l_{VGG}^{SR}= \frac{1}{W_{(i,j)}H_{(i,j)}} \sum_{x=1}^{W_{(i,j)}} \sum_{y=1}^{H_{(i,j)}} \phi_{(i,j)} (I^{HR})_{(x,y)}- \phi_{(i,j)} (G_{\theta_g} (I^{LR}))_{(x,y)}^2
    \label{eq:2}
\end{equation}

\begin{equation}
    l^{SR}= l_{VGG}^{SRR} + 10^{-3} (l_{GEN}^{SR})
    \label{eq:3}
\end{equation}

\subsection{Image Classification}
\begin{figure}[h]
\includegraphics[scale=.30,height=50mm]{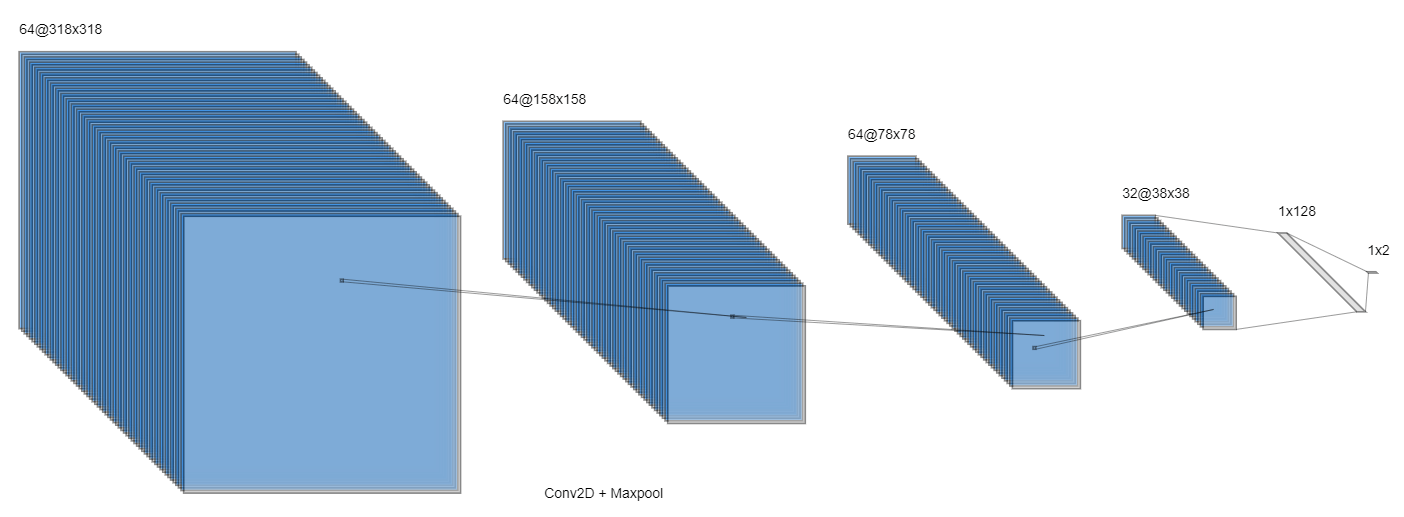}
\centering
\caption{Conv2D + Maxpool + ReLu until 38x38 latent space for both 320x320 and 80x80 input}
\label{fig:5}
\end{figure}

The image classification network (Fig \ref{fig:5}) attempts to reduce the image data to the same size latent space. This allows consistency between both networks. For both the 80px and 320px input, 2D convolutional layers are used with depth of 64, kernel size of 3 by 3, and rectified linear unit (ReLU) activation followed by 2D max pooling layers of size 2 by 2 until a feature space of 80 by 80 is achieved. We then instead use a Conv2D with depth 32 followed by a Maxpool2D with size 2 by 2 to achieve a latent space of 38 by 38. We then apply a dropout layer with rate of .25, Flatten the features, add a dense layer of size 128 with ReLU activation, a dropout layer with rate .25, and a SoftMax output to 2 classes. Loss is categorical cross-entropy and the optimizer is the default Keras Adadelta. 

\subsection{Object Detection}
The object detection framework is Mask R-CNN released by He et al. in March of 2017 \cite{he2017mask}. We used a fork of Matterport's code  \cite{matterport_maskrcnn_2017} which converted He's architecture to Python 3 and Tensorflow. We apply transfer learning from the MS COCO \cite{lin2014microsoft} pretrained Mask R-CNN model to our testing data. "Mask R-CNN is a simple, flexible, and general framework for object instance segmentation that efficiently detects objects and generates a high-quality segmentation mask. The method extends Faster R-CNN by adding a branch for predicting an object mask in parallel with the existing branch for bounding box recognition." \cite{he2017mask}

% The object detection framework is Ultralytics’ xView YoloV3 \cite{matterport_maskrcnn_2017} network. We ran the xView 2018 Object Detection challenge docker image from Ultralytics’, which provides a pre-trained PyTorch model (xview-best.pt). YoloV3 \cite{he2017mask} is an object detection model that is trained on image classes in addition to bounding boxes allowing fast object detection in one neural network.  

\section{Datasets}
\label{sec:headings}
The training set was from Skysat’s 0.8m samples and the testing set was Planetscope’s 3.0m Shipsnet images.

\subsection{Training Set}
\begin{figure}[h]
\includegraphics[scale=.475]{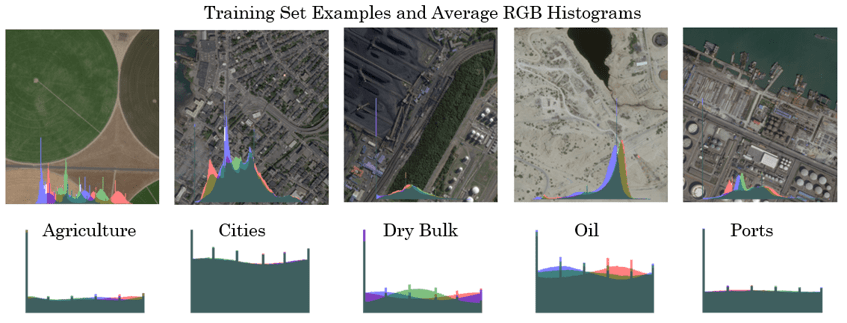}
\centering
\caption{Plant’s Skysat Visual Imagery. Example image and RGB histograms of land use classes}
\label{fig:6}
\end{figure}

To avoid having too many camera artifact variations in the dataset we want to use a single telescope’s pictures for all the training images. To this end we went with Planet’s 0.8m Visual Skysat samples \cite{skysat_samples}. This gave us the same telescope imaging system on vastly different landscapes. A visualization and RGB histogram of a single image and an average of the of each of the 5 training sets: Agriculture, Cities, Dry Bulk, Oil, and Ports, is show here (Fig \ref{fig:6}). 

Images are pansharpened \cite{padwick2010worldview}, orthorectified \cite{satellite_octo}, color corrected RGB \cite{skysat_specification_2018} and of average size 2560px by 1080px. These preprocessing steps alter the image to make it more visually appealing and useful to the user. Comparing images that have had different preprocessing would have added another variable to the experiment. Each training set contains 500 image chips with some sets reduced or slightly altered (rotation) to get to 500 images. 

\begin{wrapfigure}[10]{r}{0.3\textwidth}
    \vspace{-50pt}
    \begin{center}
        \includegraphics[width=0.25\textwidth]{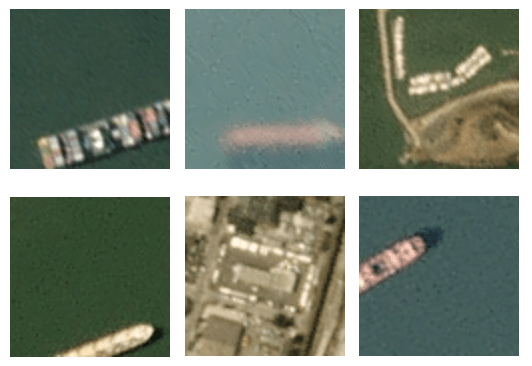}
    \end{center}
    \caption{Shipsnet Images}
    \vspace{50pt}
    \label{fig:7}
\end{wrapfigure}

\subsection{Testing Set}
The testing set contains images from the Kaggle Shipsnet competition. Example image chips are shown here (Fig \ref{fig:7}). These images are from Planet’s Planetscope satellite (3.0m) They contain images of San Francisco Bay and San Pedro Bay areas of California. Like the training sets’ images, these images are also scenes in Planet’s visual classification meaning that they are also pansharpened, orthorectified, and color corrected RGB. However, unlike the training set, the scenes in the testing set have already been tiled into 80px by 80px squares. In terms of the classes in the dataset, only full frame ships are being classified as ship. All other images are classified as no-ship. The testing set contains 4000 images with 1000 images classified as ship.

\begin{wrapfigure}[10]{r}{0.3\textwidth}
    \vspace{-90pt}
    \begin{center}
        \includegraphics[width=0.3\textwidth]{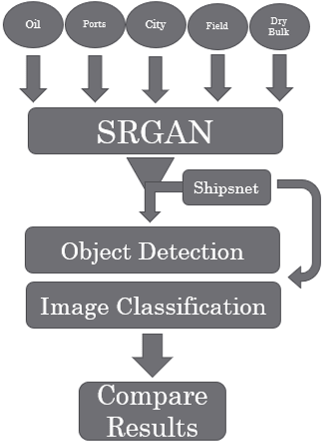}
    \end{center}
    \caption{Experimental flow chart}
    \vspace{90pt}
    \label{fig:8}
\end{wrapfigure}

\section{Experiment}
\label{sec:headings}
An overview of the experiment can be found in this flow chart (Fig \ref{fig:8}). We took the HR imagery from the training sets and down sampled them to 80 px by 80 px with a bicubic filter and then trained the SRGAN to up-sample 4x to 320x320 resolution. After the 5 SRGANs were trained we took the testing set and ran it through each of the SRGANs. We also scaled the testing set to the same 320x320 resolution to compare to the SR images directly. We then ran the SR ships, the raw ships, and the scaled ships through the image classification and object detection models to compare the performance of each computer vision task.\\

\begin{wrapfigure}[11]{r}{0.6\textwidth}
\vspace{-30pt}
    \begin{center}
        \includegraphics[width=0.60\textwidth]{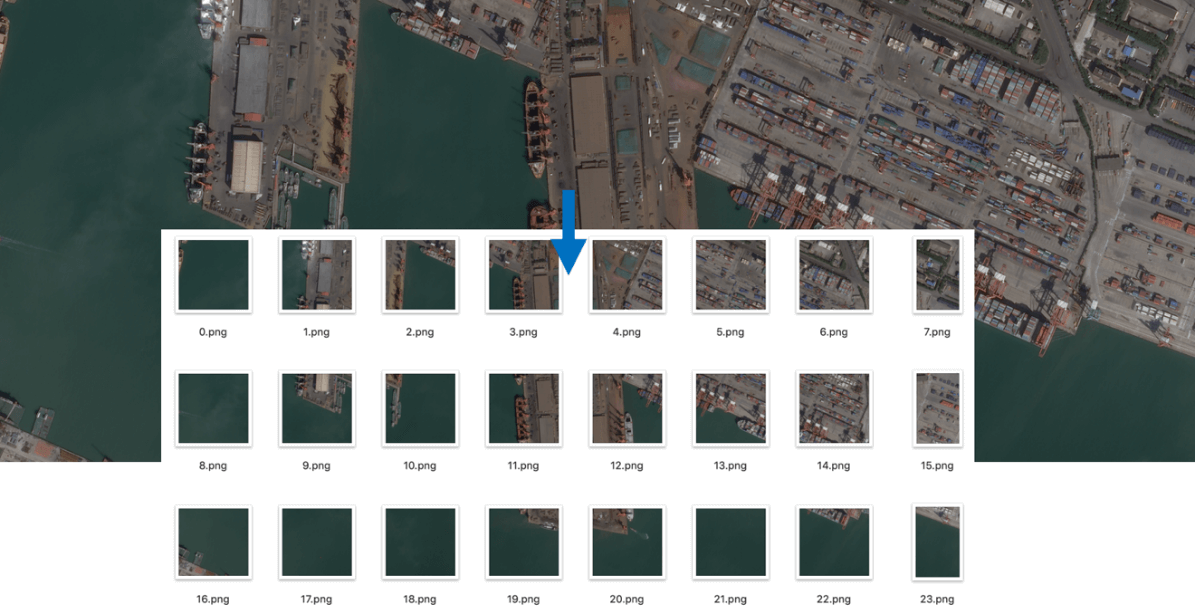}
    \end{center}
    \caption{Process of tiling image scenes into desired resolution squares}
\vspace{-30pt}
\label{fig:9}
\end{wrapfigure}

\subsection{Preprocessing}
Preprocessing the images in the training set required a few steps. Since the size of the test images are 80px by 80px we needed to get 320px by 320px images from our training set in order to satisfy the 4x up-sample. With the average size of the training set scenes being 2560px by 1080px we needed to tile the images into 320px by 320px PNG squares (Fig \ref{fig:9}). To compare the super resolution images with those of a scaled-up test set, we used Image Magick’s Mitchel-Netravali bicubic filter \cite{resampling_examples} to scale the 80px test images to 320px.

%Preprocessing the images in the training set required a few steps. Since the size of the test images are 80px by 80px we needed to get 320px by 320px images from our training set in order to satisfy the 4x up-sample. With the average size of the training set scenes being 2560px by 1080px we needed to tile the TIFF images into 320px by 320px PNG squares (Fig \ref{fig:9}). This process was also done on the testing sets for the SRGAN satellite image system comparison. To compare the super resolution images with those of a scaled-up test set, we used Image Magick’s Mitchel-Netravali filter \cite{resampling_examples}. This is a bicubic filter with parameters B=1/3 and C=1/3. We used this filter to scale the 80px by 80px testing set to 320px by 320px so we could compare them with the SR images.

\subsection{SRGAN}
With all our images in the required resolution we can now setup our SRGAN network. After loading our data, we need to first down sample to get our LR image. We again use a bicubic filter to down sample the images to reduce the effect of the sampling method on the results as talked about in \cite{kawulok2019training} by Kawulok et al. We then mean normalize all pixel values between -1 and 1 before running the data through the GAN. We train 5 SRGAN’s, one for each training set, before moving on to the computer vision tasks. We trained for 5000 epochs with a batch size of 16 for 500 images. After training the SRGANs, we ran inference on some down-samples of popular satellite image samples: Xview(.3m), Pleides(.5m), Quickbird(.65m), Triplesat(.8m), and Ikonos(1m). In addition, PSNR/SSIM metrics were calculated for each SR model on each of the training datasets to compare how a SR model that was trained on its own test data performed against the other SR models.

%With all our images in the required resolution we can now setup our SRGAN network. After loading our data, we need to first down sample to get our LR image. To do so we use SciPy’s ‘bicubic’ interpolation over a 4px by 4px area. Bicubic interpolation was used in both up sampling and down sampling to reduce the effect of the sampling method on the results as talked about in \cite{kawulok2019training} by Kawulok et al. We then mean normalize all pixel values between -1 and 1 before running the data through the GAN. We train 5 SRGAN’s, one for each training set, before moving on to the computer vision tasks. We trained for 5000 epochs with a batch size of 16 for 500 images. After training the SRGANs, we ran inference on some down-samples of popular satellite image samples: Xview(.3m), Pleides(.5m), Quickbird(.65m), Triplesat(.8m), and Ikonos(1m).

\subsection{Image Classification}
The image classification step is straight forward. We create two networks to accommodate the 320px and 80px inputs. As described in the network details, we run the data through convolutional layers and max pools until we reach the same 38x38 latent space size. After creating the network, we trained the image classifier on our 7 data sources: the 5 SR images, the scaled images, and the raw images. We use a batch size of 32, 100 epochs, 20\% validation split and normalize data between 0 and 1. In addition we apply some data augmentation: 10-degree random rotation, random width shift and height shift of 0.1, and random horizontal flip. 

\subsection{Object Detection}
To run the image through the object detection framework we need to first grab the images from SRGAN and annotate them. To label for object detection we needed to create PASCAL Visual Object Classes (VOC) object detection xml files. We used LabelImg \cite{labelimg} to draw bounding boxes on our 1000 ships for the SR images and the raw images (all the SR  images have the same annotations). Mask R-CNN is built on FPN \cite{lin2016feature} and ResNet101 \cite{he2015deep}. We used the default parameters as provided by the config file of Matterport's repo \cite{matterport_maskrcnn_2017} while running 5 epochs. We take the 1000 labeled images and split it into 700 training images, 150 testing images and 150 validation images.

% To run the images through the object detector we decided to create a montage of each set's images (Fig \ref{fig:10}) to easily run and measure the detection rates. Creating the montage was done with Image Magick with a border pixel size of 1 and image shape of 25 by 40 for the 1000 images. The final width and height for each of the 6 montage images was 12960 px by 8100 px and the size of the raw montage at 3360 px by 2100 px. Each montage then underwent inference of the pretrained xview yolov3 pytorch model outputting a bounding box tagged image. The detection script was run with img-size of 1632, object confidence threshold of .99, and the xview\_best\_lite.pt model. We then calculated precision and other classification metrics.

% To run the images through the object detector we decided to montage each data sources’ images into one image (Fig \ref{fig:10}) to easily run and measure the detection rates. This montaging was done with Image Magick with a border pixel size of 1 and image shape of 25 by 40 for the 1000 images. The final width and height for each of the 6 montage cases was 12960 px by 8100 px at 24-bit depth and the size of the raw montage at 3360 px by 2100 px. Each montage then underwent inference outputting a bounding box tagged image and detection scores.

\section{Results}
\label{sec:headings}
The experiment took approximately a week from start to finish for just training. Training for each SRGAN took around 48 GPU hours each on a Nvidia V100 32GB. Each model was trained on a single GPU. Inference for SRGAN was around 37 fps for the 17MB model on a Xeon E5-2698 v4 (50M Cache, 2.20 GHz). Image classification models took around 2 hours to train while object detection models took 20 minutes on a Nvidia V100.

\pagebreak 

\subsection{SRGAN}
\begin{figure}[H]
    \centering
    \includegraphics[width=.95\textwidth]{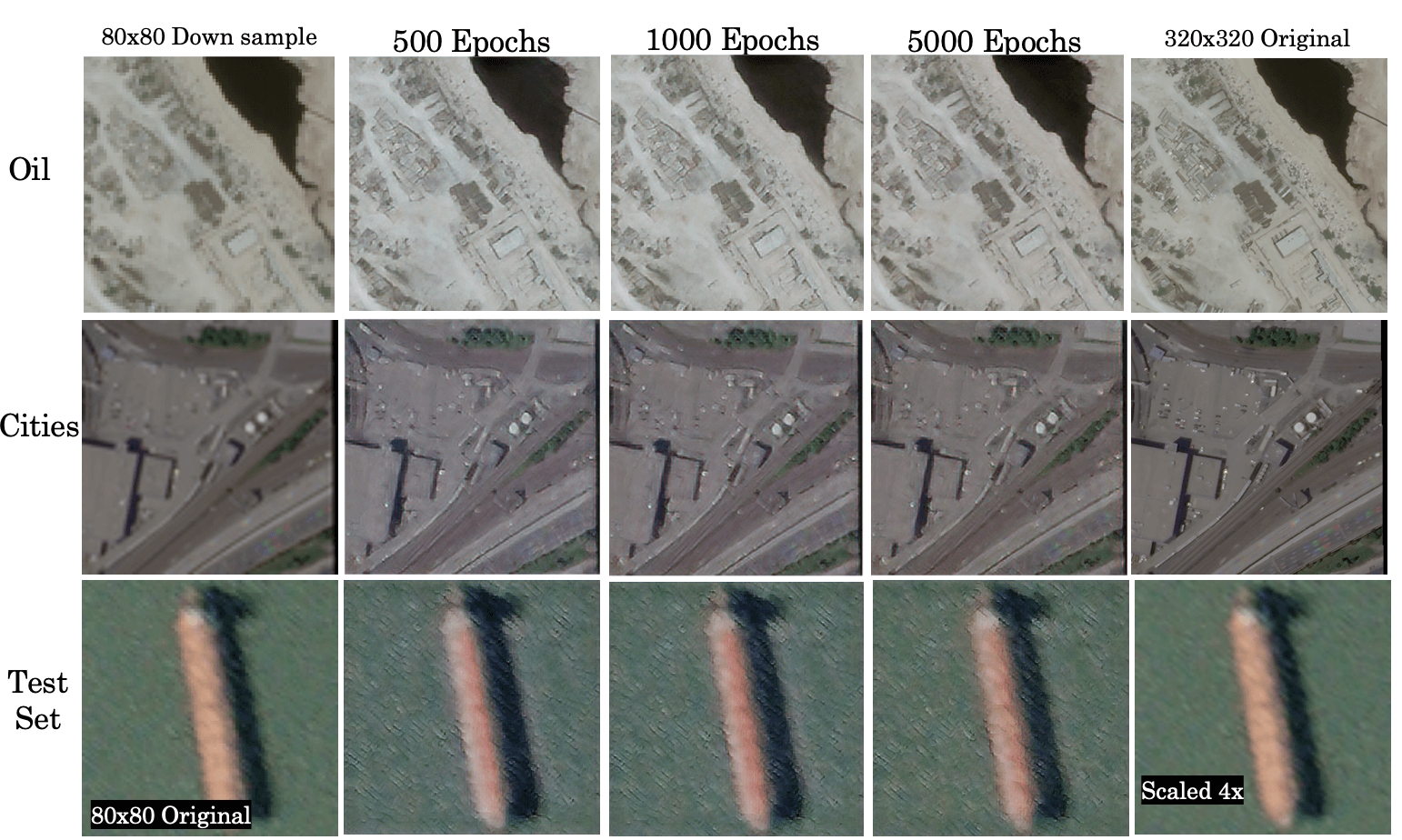}
    \caption{Improvement of SRGANs over epochs}
    \label{fig:11}
\end{figure}

\begin{figure}[H]
    \includegraphics[width=.95\textwidth]{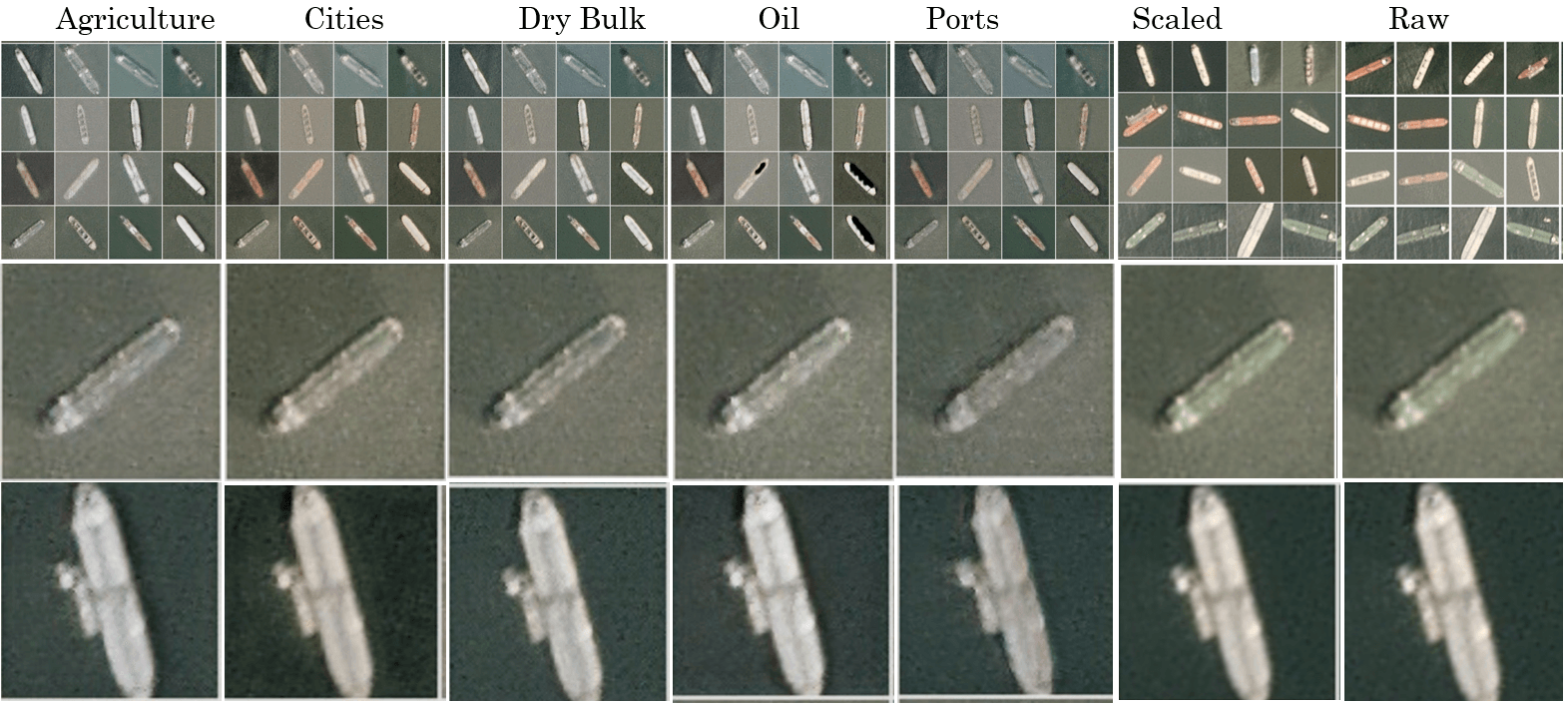}
    \centering
    \caption{Montage image compare between the 7 testing sets}
    \label{fig:19}
\end{figure}

We can see the progression over the 5000 epochs in (Fig \ref{fig:11}). We ran a couple SR training image in addition to running inference on the test set. We can see that over time the model learned a better color mapping between the super resolved images and the original images. In addition, over time a more defined building structure can be seen in the desert picture highlighting the strength of the perceptual loss function.\\
\phantom{x}\hspace{3ex} For images of the ships (Fig \ref{fig:19}), the Agriculture trained SRGAN produced a much darker image and the Cities trained SRGAN produced a much greyer image. This could possibly be due to the different color distributions present in the training sets. We also notice the artifacting present in some of the SR images, particular the "black spots" located in the Oil trained SRGAN. 

% \begin{figure}[H]
%   \centering
%   \begin{minipage}[b]{0.64\textwidth}
%     \includegraphics[width=\textwidth]{Figures/F6.png}
%     \caption{Improvement of SRGANs over epochs}
%     \label{fig:11}
%   \end{minipage}
%   \hfill
%   \begin{minipage}[b]{0.3475\textwidth}
%     \includegraphics[width=\textwidth]{Figures/F7.png}
%     \caption{SRGANs on the same image}
%     \label{fig:12}
%   \end{minipage}
% \end{figure}

\begin{figure}[H]
    \centering
    \includegraphics[width=\textwidth]{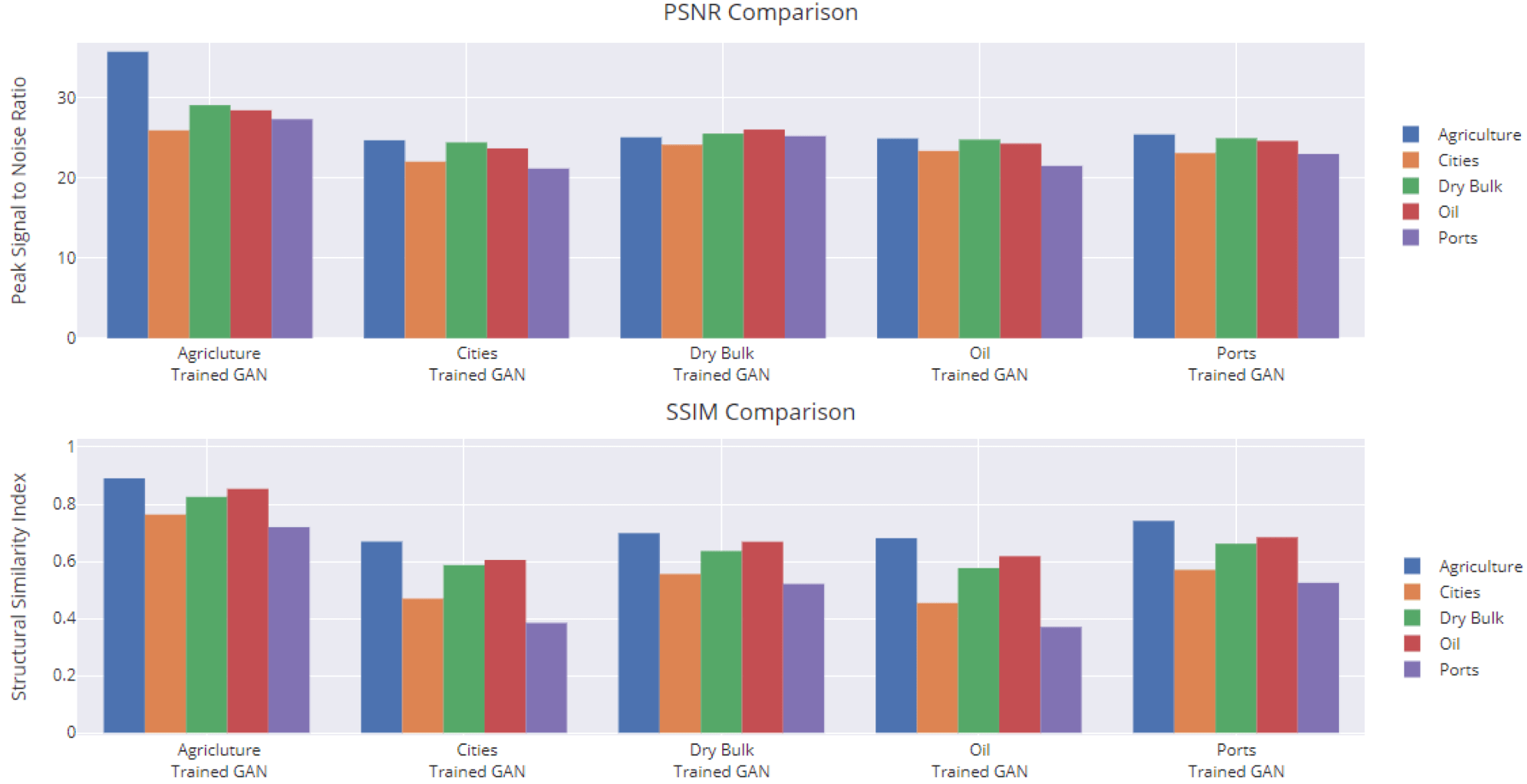}
    \caption{PSNR/SSIM metrics for SR models with all training sets}
    \label{fig:22}
\end{figure}

\begin{figure}[H]
    \centering
    \includegraphics[width=\textwidth]{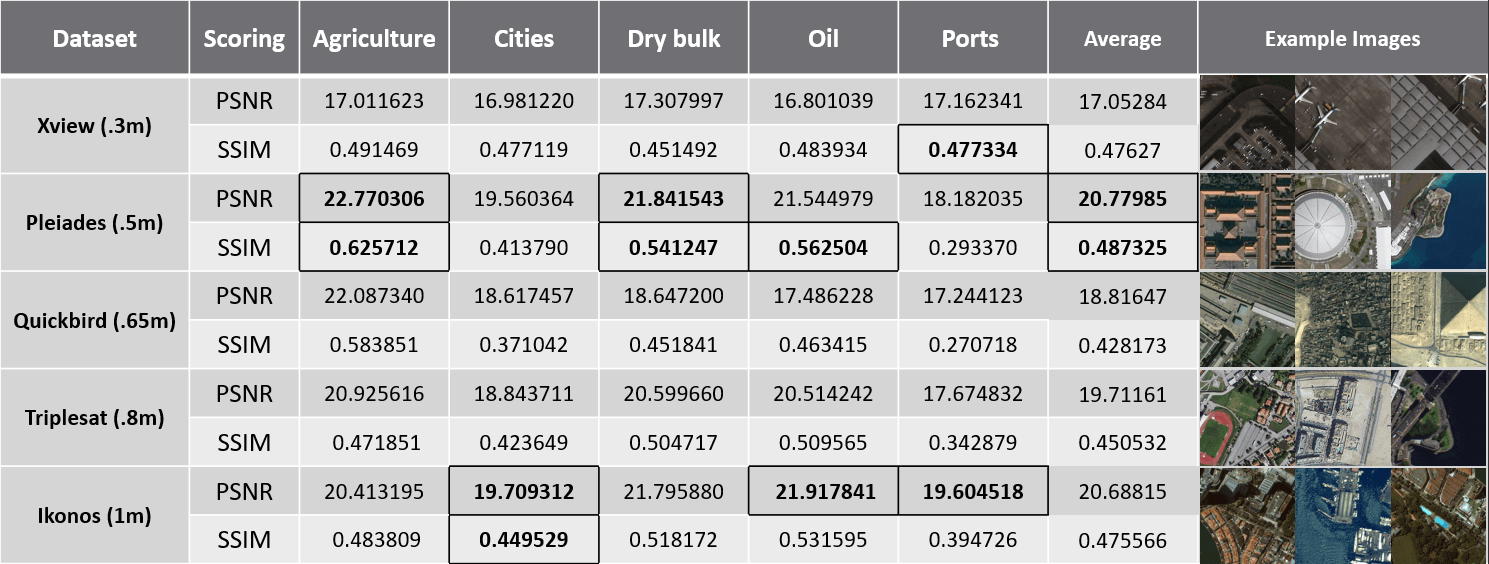}
    \caption{Scores for SRGAN applied to sample images from different satellites. Averages for 50-200 image chips.}
    \label{fig:13}
\end{figure}

The comparison (Fig \ref{fig:22}) between SR models and all the training sets show consistent relative performance for each model across all the datasets. We see that the models did the best on the homogeneous agriculture images while they did the worst on the images of heterogeneous cities. \\
\phantom{x}\hspace{3ex} In terms of metrics, image quality scores for various satellite image samples are shown here (Fig \ref{fig:13}). To show SRGANs sensitivity to spatial resolution of the training set we tested various 4x down-samples. For example Pleiades' .5m spatial resolution was down-sampled 4x to 2m  while Triplesat's .8m spatial resolution was down-sampled to 3.2m before being super resolved and compared to the original image. This is meant to highlight the need for a tailored training set on not only the ontology but also the spatial resolution.

\pagebreak 

\subsection{Image Classification}
\begin{figure}[H]
    \includegraphics[scale=.375]{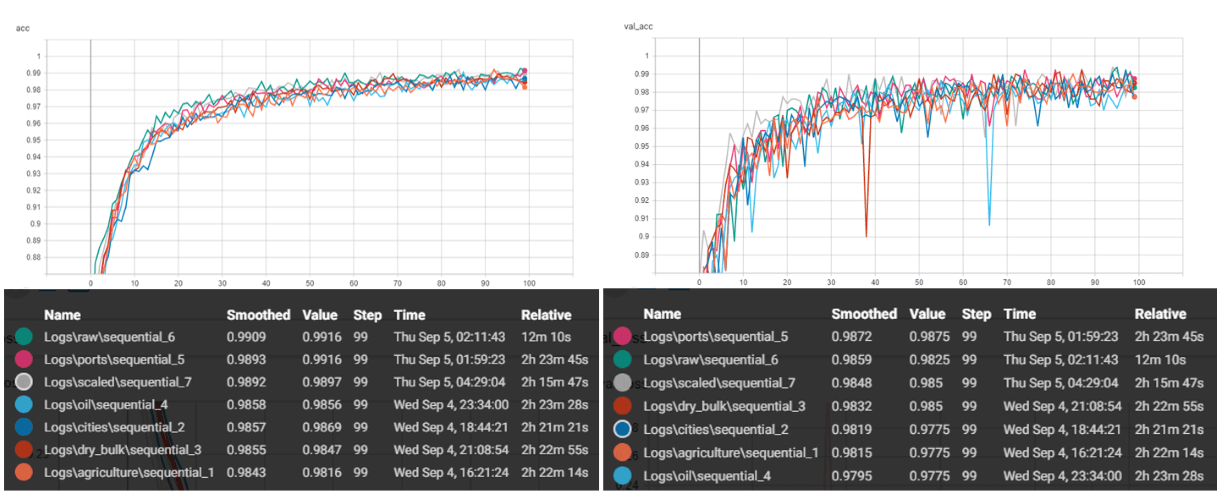}
    \caption{Training and Validation accuracy for the image classification network using the Super Resolution images, the raw images, and the scaled images}
    \label{fig:14}
\end{figure}

\begin{figure}[H]
    \includegraphics[scale=.39]{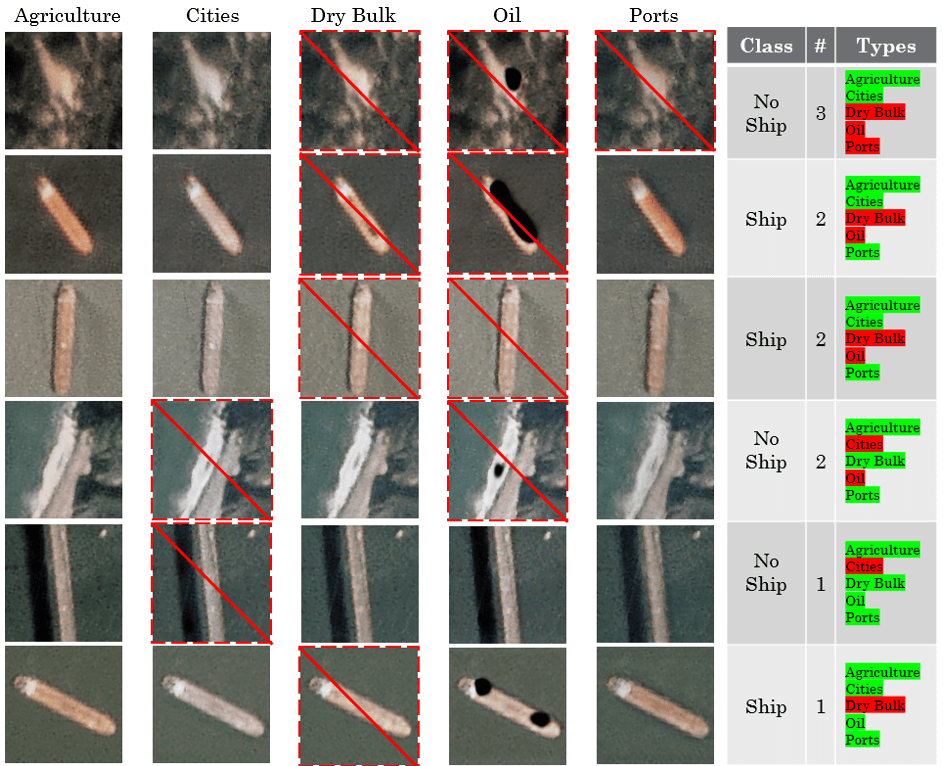}
    \centering
    \caption{Image classifications failure highlights}
    \label{fig:16}
\end{figure}

Accuracy over epochs shown here (Fig \ref{fig:14}). Overall, image classification went very smoothly due to the already high validation accuracy on the raw dataset (98\%). However, we can look at the few misclassifications for our networks (Fig \ref{fig:16}). Results show that the SRGANs that made the most artifacts (i.e. oil trained SRGAN), had lower image classification scores because of it. We see that the ports SR images outperformed the raw validation accuracy by 20 images.

\pagebreak 

\subsection{Object Detection}
\begin{figure}[H]
    \includegraphics[width=\textwidth]{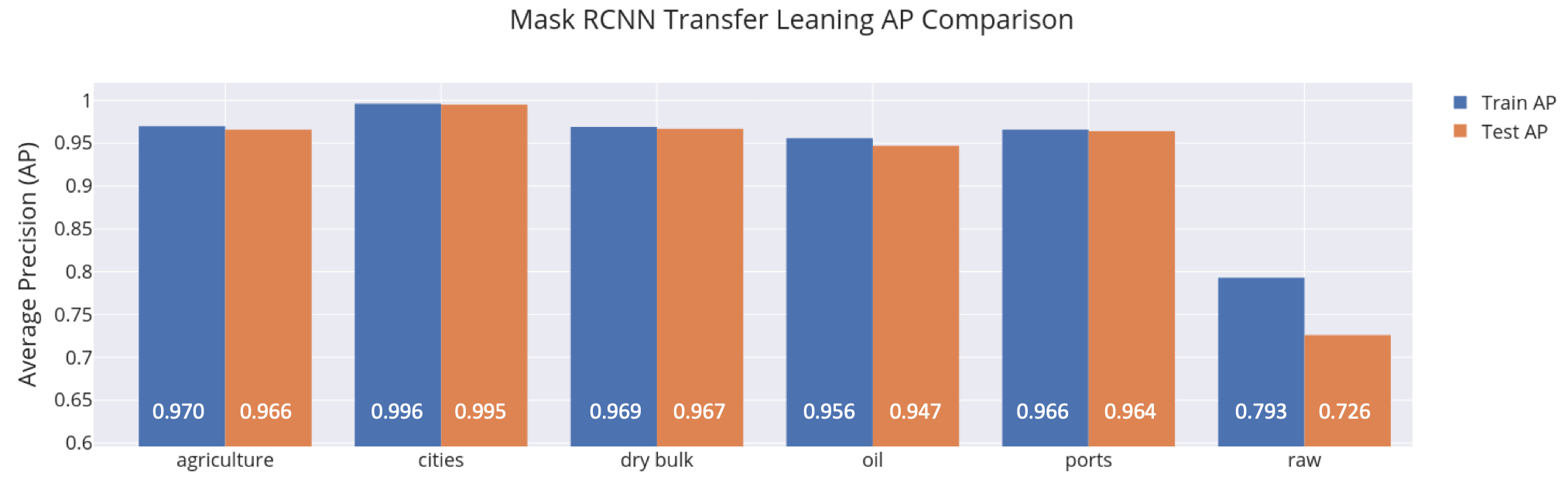}
    \centering
    \caption{Train/Test Average Precision}
    \label{fig:20}
\end{figure}

\begin{figure}[H]
    \includegraphics[width=0.95\textwidth]{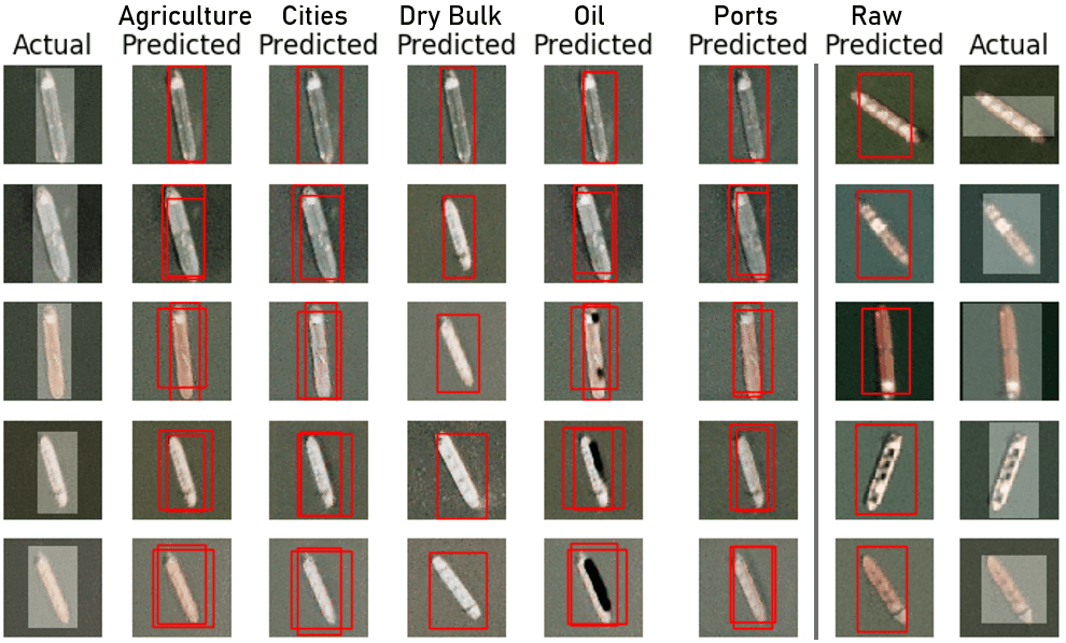}
    \centering
    \caption{Mask RCNN Object Detection Comparisons}
    \label{fig:21}
\end{figure}

The results for object detection show a clear improvement in average precision (AP) when comparing the SR images to the raw images trained Mask R-CNN (Fig \ref{fig:20}). AP for both the training set and the test set are provided. We notice that the cities SR images have a near perfect precision with missing only 4 ships in our training set and 1 image in the testing set.  On average, the SR models had a 18.4\% higher AP than the raw image trained Mask R-CNN.\\
\phantom{x}\hspace{3ex} We can see the actual predictions made by the network in (Fig \ref{fig:21}).We can see the multiple detections made on the same ship possible due to a lower than needed confidence threshold but overall the models performed very well when compared to the raw image trained network. 

% The results for object detection show a clear improvement in average precision (AP) when comparing the SR images to the raw image (Fig \ref{fig:19}) Examples of object detection are found here (Fig \ref{fig:19}). Cities trained SRGAN have the greyest outcome, perhaps revealing more detail in urban environments. Ports trained SRGAN have maritime-like borders but do not noticeably outperform training data without shorelines or mixed land-water. Oil trained SRGAN generates the highest false positive rate despite having a lot of artifacts. Scaled images generate poorly sized bounding boxes relative to 0.3 m GSD for xView YOLOv3 baseline case. \\
% \phantom{x}\hspace{3ex} The fixed resolution of overhead imagery doesn’t seem to benefit much by augmentation and shows sensitivity to resolution changes based on size. An example of the xView’s sensitivity to scale is at a higher resolution, the same image can detect a building mistakenly for a car, and at lower resolution, a car can be mistaken as a yacht.  Overall, for the base model on unaltered xView imagery, the best validation mAP is 0.16 after 300 epochs (3 days), corresponding to a training mAP of 0.30.

\pagebreak 

\section{Conclusion}
\label{sec:headings}
Clear differences are seen between the SRGAN networks when trained on different sets of imagery. Performance of our super resolution (4x) is similar to the state of the art for non-satellite imagery (Agriculture vs. Urban 100) standing at PSNR / SSIM scores around 27.7 / .79 vs. 27.1 / .82. For the SR models with all the training sets comparison we notice consistent relative strength. This indicates that the quality of the data that a SR model is trained on is paramount.\\
\phantom{x}\hspace{3ex} Image classification showed preference for being trained on images that were in the testing set with marginally higher accuracy for SRGANs trained on images of ships. Only 1 (ports) of the 5 trained SRGANs had higher validation accuracy (98.72\%) than the raw test set (98.59\%). Depending on the difficulty of the test set, will see much smaller, if any, increase in image classification performance if raw accuracy is already high.\\
\phantom{x}\hspace{3ex} For object detection, we see an average of 18.4\% improvement in precision when comparing the SR trained model versus the raw image trained model. The cities trained Mask R-CNN followed by the ports trained model preformed the best, out recognizing around 100 images when compared to the raw image trained model. This suggest that having objects in the test regime (ports) and having a complex and diverse set of images (cities) gives good results for super resolution for object detection.\\
\phantom{x}\hspace{3ex} Overall, we notice a clear trend in that having a diverse dataset with objects in the test ontology allow downstream tasks to perform better after super resolution. 

\subsection{Next Steps}
Future work with testing on various datasets \cite{takano2019srgan} will validate the increase in performance from SR networks. Using neural networks that allow varying image size inputs would allow one to use SR on images of arbitrary size unlike the fixed size of SRGAN. Many frameworks are currently out there that do this \cite{idealo_2019}. These networks avoid the user’s need to preprocess the images into a required format since they handle various sizes. Refining the method used for object detection could yield clearer results. Future testing, with different neural network architectures \cite{dai2019second}, could yield better results. An entire view into super resolution can be found here \cite{anwar2019deep}.

\subsection{Extras}
\begin{figure}[H]
    \centering
    \includegraphics[width=\textwidth]{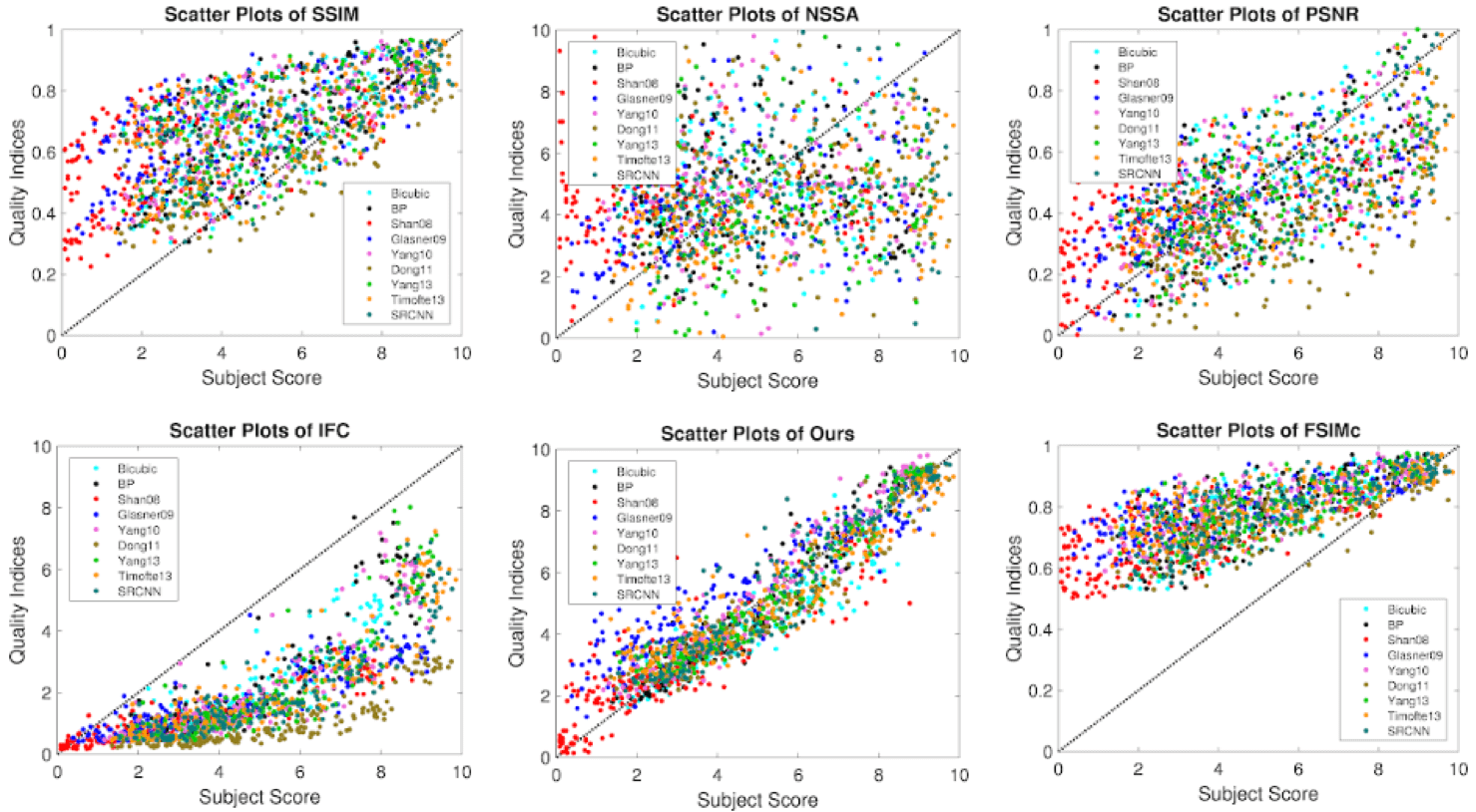}
    \caption{Ma's \cite{ma2017learning} Human Subject Scores vs Image Metrics}
    \label{fig:23}
\end{figure}

An interesting paper we came across in our research was Chao Ma's et al. \cite{ma2017learning} with their comparison of human subject scores as compared with image metrics (Fig \ref{fig:23}). This shows that the commonly used metrics are not always as telling of the quality of an image. 

\begin{figure}[H]
    \centering
    \includegraphics[width=\textwidth]{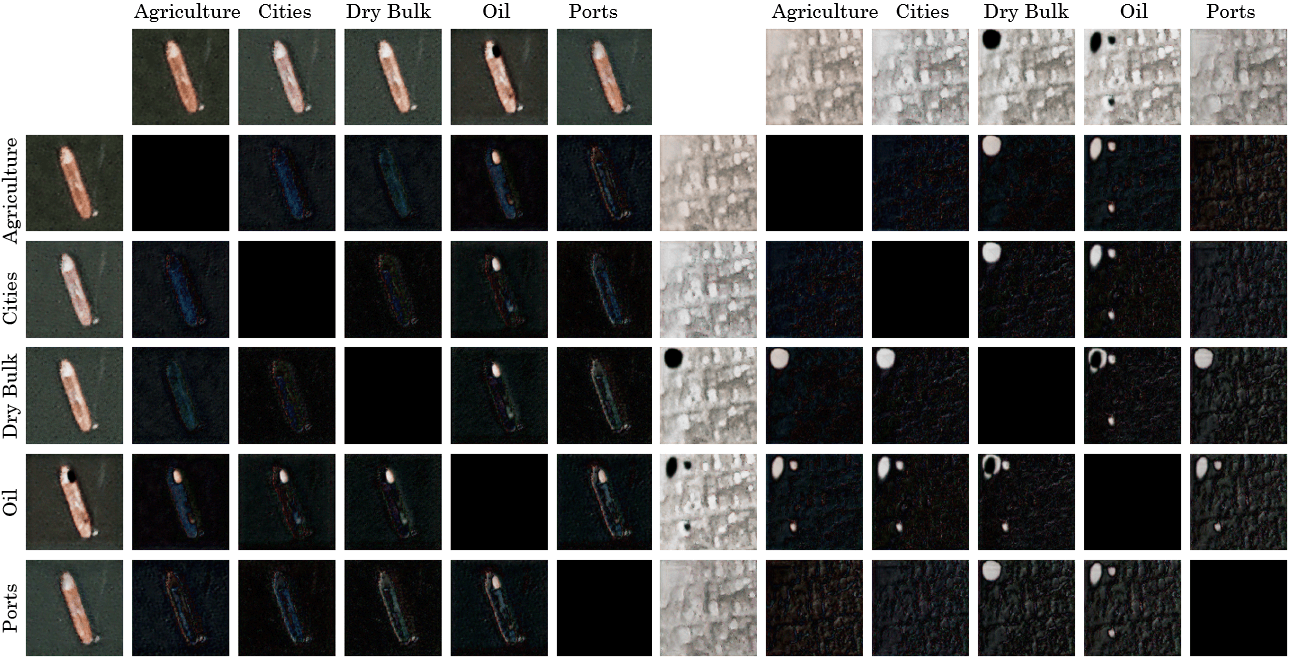}
    \caption{Image comparison between SRGANs}
    \label{fig:24}
\end{figure}

To show the difference between SR models we ran an absolute difference matrix with two sample images (Fig \ref{fig:24}).This comparison really emphasized the artifacting that each SRGAN model learned during training. For example, clear artifacting "black spots" are seen in the dry bulk SR image of the city.

\bibliographystyle{unsrt}
\bibliography{references}

\begin{thebibliography}{10}

\bibitem{yang2019deep}
Wenming Yang, Xuechen Zhang, Yapeng Tian, Wei Wang, Jing-Hao Xue, and Qingmin
  Liao.
\newblock Deep learning for single image super-resolution: A brief review.
\newblock {\em IEEE Transactions on Multimedia}, 2019.

\bibitem{keys1981cubic}
Robert Keys.
\newblock Cubic convolution interpolation for digital image processing.
\newblock {\em IEEE transactions on acoustics, speech, and signal processing},
  29(6):1153--1160, 1981.

\bibitem{dai2009softcuts}
Shengyang Dai, Mei Han, Wei Xu, Ying Wu, Yihong Gong, and Aggelos~K
  Katsaggelos.
\newblock Softcuts: a soft edge smoothness prior for color image
  super-resolution.
\newblock {\em IEEE Transactions on Image Processing}, 18(5):969--981, 2009.

\bibitem{kawulok2019training}
Michal Kawulok, Szymon Piechaczek, Krzysztof Hrynczenko, Pawel Benecki, Daniel
  Kostrzewa, and Jakub Nalepa.
\newblock On training deep networks for satellite image super-resolution.
\newblock {\em arXiv preprint arXiv:1906.06697}, 2019.

\bibitem{agustsson2017ntire}
Eirikur Agustsson and Radu Timofte.
\newblock Ntire 2017 challenge on single image super-resolution: Dataset and
  study.
\newblock In {\em Proceedings of the IEEE Conference on Computer Vision and
  Pattern Recognition Workshops}, pages 126--135, 2017.

\bibitem{sentinel2a}
Sentinel-2a (10m) satellite sensor.

\bibitem{shermeyer2019effects}
Jacob Shermeyer and Adam Van~Etten.
\newblock The effects of super-resolution on object detection performance in
  satellite imagery.
\newblock In {\em Proceedings of the IEEE Conference on Computer Vision and
  Pattern Recognition Workshops}, pages 0--0, 2019.

\bibitem{kim2016accurate}
Jiwon Kim, Jung Kwon~Lee, and Kyoung Mu~Lee.
\newblock Accurate image super-resolution using very deep convolutional
  networks.
\newblock In {\em Proceedings of the IEEE conference on computer vision and
  pattern recognition}, pages 1646--1654, 2016.

\bibitem{van2018you}
Adam Van~Etten.
\newblock You only look twice: Rapid multi-scale object detection in satellite
  imagery.
\newblock {\em arXiv preprint arXiv:1805.09512}, 2018.

\bibitem{liu2016ssd}
Wei Liu, Dragomir Anguelov, Dumitru Erhan, Christian Szegedy, Scott Reed,
  Cheng-Yang Fu, and Alexander~C Berg.
\newblock Ssd: Single shot multibox detector.
\newblock In {\em European conference on computer vision}, pages 21--37.
  Springer, 2016.

\bibitem{takano2019srgan}
Nao Takano and Gita Alaghband.
\newblock Srgan: Training dataset matters.
\newblock {\em arXiv preprint arXiv:1903.09922}, 2019.

\bibitem{liu2018large}
Ziwei Liu, Ping Luo, Xiaogang Wang, and Xiaoou Tang.
\newblock Large-scale celebfaces attributes (celeba) dataset.
\newblock {\em Retrieved August}, 15:2018, 2018.

\bibitem{yu2015lsun}
Fisher Yu, Ari Seff, Yinda Zhang, Shuran Song, Thomas Funkhouser, and Jianxiong
  Xiao.
\newblock Lsun: Construction of a large-scale image dataset using deep learning
  with humans in the loop.
\newblock {\em arXiv preprint arXiv:1506.03365}, 2015.

\bibitem{ledig2017photo}
Christian Ledig, Lucas Theis, Ferenc Husz{\'a}r, Jose Caballero, Andrew
  Cunningham, Alejandro Acosta, Andrew Aitken, Alykhan Tejani, Johannes Totz,
  Zehan Wang, et~al.
\newblock Photo-realistic single image super-resolution using a generative
  adversarial network.
\newblock In {\em Proceedings of the IEEE conference on computer vision and
  pattern recognition}, pages 4681--4690, 2017.

\bibitem{deepak112_2019}
deepak112.
\newblock deepak112/keras-srgan, Apr 2019.

\bibitem{johnson2016perceptual}
Justin Johnson, Alexandre Alahi, and Li~Fei-Fei.
\newblock Perceptual losses for real-time style transfer and super-resolution.
\newblock In {\em European conference on computer vision}, pages 694--711.
  Springer, 2016.

\bibitem{he2017mask}
Kaiming He, Georgia Gkioxari, Piotr Doll{\'a}r, and Ross Girshick.
\newblock Mask r-cnn.
\newblock In {\em Proceedings of the IEEE international conference on computer
  vision}, pages 2961--2969, 2017.

\bibitem{matterport_maskrcnn_2017}
Waleed Abdulla.
\newblock Mask r-cnn for object detection and instance segmentation on keras
  and tensorflow.
\newblock \url{https://github.com/matterport/Mask_RCNN}, 2017.

\bibitem{lin2014microsoft}
Tsung-Yi Lin, Michael Maire, Serge Belongie, James Hays, Pietro Perona, Deva
  Ramanan, Piotr Doll{\'a}r, and C~Lawrence Zitnick.
\newblock Microsoft coco: Common objects in context.
\newblock In {\em European conference on computer vision}, pages 740--755.
  Springer, 2014.

\bibitem{skysat_samples}
Skysat sample imagery, 2019.

\bibitem{padwick2010worldview}
Chris Padwick, Michael Deskevich, Fabio Pacifici, and Scott Smallwood.
\newblock Worldview-2 pan-sharpening.
\newblock In {\em Proceedings of the ASPRS 2010 Annual Conference, San Diego,
  CA, USA}, volume 2630, 2010.

\bibitem{satellite_octo}
Satellite~Imaging Corporation.
\newblock Orthorectification, 2019.

\bibitem{skysat_specification_2018}
Planet.
\newblock Skysat imagery products specification, Nov 2018.

\bibitem{resampling_examples}
Imagemagick v6 examples -- resampling filters.

\bibitem{labelimg}
tzutalin.
\newblock Labelimg is a graphical image annotation tool and label object
  bounding boxes in images.
\newblock \url{https://github.com/tzutalin/labelImg}, 2017.

\bibitem{lin2016feature}
T~Lin, P~Doll{\'a}r, RB~Girshick, K~He, B~Hariharan, and SJ~Belongie.
\newblock Feature pyramid networks for object detection. corr, vol.
\newblock {\em arXiv preprint arXiv:1612.03144}, 2016.

\bibitem{he2015deep}
Kaiming He, Xiangyu Zhang, Shaoqing Ren, and Jian Sun.
\newblock Deep residual learning for image recognition. corr abs/1512.03385
  (2015), 2015.

\bibitem{idealo_2019}
Idealo.
\newblock idealo/image-super-resolution, Sep 2019.

\bibitem{dai2019second}
Tao Dai, Jianrui Cai, Yongbing Zhang, Shu-Tao Xia, and Lei Zhang.
\newblock Second-order attention network for single image super-resolution.
\newblock In {\em Proceedings of the IEEE Conference on Computer Vision and
  Pattern Recognition}, pages 11065--11074, 2019.

\bibitem{anwar2019deep}
Saeed Anwar, Salman Khan, and Nick Barnes.
\newblock A deep journey into super-resolution: A survey.
\newblock {\em arXiv preprint arXiv:1904.07523}, 2019.

\bibitem{ma2017learning}
Chao Ma, Chih-Yuan Yang, Xiaokang Yang, and Ming-Hsuan Yang.
\newblock Learning a no-reference quality metric for single-image
  super-resolution.
\newblock {\em Computer Vision and Image Understanding}, 158:1--16, 2017.

\end{thebibliography}

\end{document}